\pdfoutput=1

\documentclass[11pt]{article}

\usepackage[]{EMNLP2023}

\usepackage{times}
\usepackage{latexsym}

\usepackage[T1]{fontenc}

\usepackage[utf8]{inputenc}

\usepackage{microtype}

\usepackage{inconsolata}

\usepackage{booktabs}

\title{mLongT5: A Multilingual and Efficient Text-To-Text Transformer for Longer Sequences}

\author{David Uthus, Santiago Onta\~{n}\'{o}n, Joshua Ainslie, Mandy Guo \\
  {\rm Google Research}\\
 \{duthus, santiontanon, jainslie, xyguo\}@google.com
}

\begin{document}
\maketitle
\begin{abstract}
We present our work on developing a multilingual, efficient text-to-text transformer that is suitable for handling long inputs.
This model, called mLongT5, builds upon the architecture of LongT5, while leveraging the multilingual datasets used for pretraining mT5 and the pretraining tasks of UL2.
We evaluate this model on a variety of multilingual summarization and question-answering tasks, and the results show stronger performance for mLongT5 when compared to existing multilingual models such as mBART or M-BERT.
\end{abstract}

\section{Introduction}

In recent years, there has been development of making transformer-based models more efficient so that they can handle longer input sequences.
Many of the models though have been English-only, making them inapplicable to other languages.

In this paper, we present our work in extending one of these models to be able to handle multilingual data.
Our model, called \emph{mLongT5}, takes advantage of the efficient architecture of LongT5 \cite{guo-etal-2022-longt5}, and has been pretrained on the multilingual mC4 dataset \cite{xue-etal-2021-mt5} to be able to work on multilingual tasks.
We have applied mLongT5 to a variety of multilingual summarization and question-answering tasks, and results show that mLongT5 exhibits strong performance in these domains.

The configurations\footnote{\url{https://github.com/google/flaxformer/tree/main/flaxformer/t5x/configs/longt5/models}} and checkpoints\footnote{\url{https://github.com/google-research/longt5}} have all been open-sourced.

\section{Related Work}

There are two areas of related work -- efficient transformer models that can handle long inputs, and multilingual models.

There has been much interest of late in making transformer models more efficient, such as to handle longer inputs.
Example of these include ETC \cite{ainslie2020etc}, Big Bird \cite{zaheer-bigbird}, LongT5 \cite{guo-etal-2022-longt5}, and Longformer \cite{beltagy-longformer}.
These models were successful in taking various approaches to address the quadratic growth of the attention mechanism in transformers.
Unfortunately though, these models are trained on English datasets, limiting their use in multilingual domains.

With respect to multilingual models, these would include mT5 \cite{xue-etal-2021-mt5}, mBART \cite{liu-mbart}, and the recent umT5 \cite{chung2023unimax}.
These models re-used architectures used by English models but are pretrained on a larger, multilingual corpus, with mT5 and umT5 trained on 101 languages and mBART on 25.
While these models showed strong performance on being able to handle a wide variety of languages, they suffered the same restrictions as their original English models on not being able to scale up to longer sequences.

\section{Model}

mLongT5 builds upon the architecture of LongT5 \cite{guo-etal-2022-longt5}.
LongT5 was developed to efficiently handle long inputs by utilizing a more efficient attention mechanism.
The model was shown to have strong performance on a variety of downstream tasks, and thus is the foundation for mLongT5.

\subsection{Datasets}

To make mLongT5 multilingual, we leverage the mC4 dataset used for training the multilingual model mT5 \cite{xue-etal-2021-mt5}, which consists of 101 languages.
This dataset has recently been updated, as described by \citet{chung2023unimax}, and was used for training umT5 and creating a new SentencePiece model \cite{kudo-richardson-2018-sentencepiece}.
As such, we then make use of the same SentencePiece model used for umT5, thus allowing mLongT5 to handle multilingual inputs.

\subsection{Pretraining Tasks}

One key difference with our model and LongT5 is the changing of tasks for pretraining the model.
LongT5 made use of PEGASUS' Principle Sentences Generation (PSG) \cite{pegasus} for pretraining its models.
While this was shown to have strong performance for various downstream tasks, the one weakness of PSG is that it is less suitable for multilingual training.
PSG relies on being able to split a piece of text into sentences, with current implementation best suited for Latin-based languages.
The need to break text into sentences properly for 101 different languages makes it then a challenging task to use in a multilingual setting.

To overcome this, we instead decided to apply UL2's pretraining tasks \cite{ul2}.
Their pretraining task, called Mixture-of-Denoisers (MoD), has the model learning from a mixture of tasks, and has been shown to work better than T5's original pretraining task \cite{t5}.
More importantly, MoD can be more easily applied to other languages compared to PSG, thus making it ideal for pretraining mLongT5.

\subsection{Pretraining Details}

Pretraining mLongT5 has many similarities to how LongT5 was pretrained.
It is pretrained for one million steps, and we pretrained model sizes of Base, Large, and XL.
We also use the same pretraining lengths, 4,096 for the inputs and 910 for the targets.
One small difference is increasing the batch size from 128 to 256, allowing the model to train on the same number of tokens as mT5.
For the mC4 dataset, we used version 3.1.0, which is the version update by \citet{chung2023unimax}.
For dataset sampling, we use the UniMax sampling method \cite{chung2023unimax}.

Instead of PSG as pretraining task, we apply MoD, using the same configuration as defined in the original UL2 task definition.
The only exception is that we do not use 0.5 corruption rate (using only corruption rate of 0.15), as our input lengths (4096) are much longer than our target lengths (910), making a corruption rate of 0.5 unfeasible.

All models were pretrained using 256 TPUv4 chips.
Wall time to pretrain these models was 1.9 days for Base, 3.7 days for Large, and 12.4 days for XL.

\section{Results}

As with the original LongT5 paper, we look at two domains for evaluating our model: summarization and question answering.

For all of these tasks, we use the default values as used for T5 finetuning, only explicitly setting the input and target lengths as described in the tasks below.

\subsection{Summarization}

The three summarization tasks we are looking at are:

\begin{itemize}
    \item MLSUM \cite{scialom-etal-2020-mlsum}: a collection of newspaper articles and their corresponding summaries in five languages: French, German, Spanish, Russian, and Turkish.
    
    \item XL-Sum \cite{hasan-etal-2021-xl}: a collection of BBC articles and summaries in 44 languages.
    
    \item WikiLingua \cite{ladhak-etal-2020-wikilingua}: a collection of documents from WikiHow (in Spanish, Turkish, Russian, and Vietnamese) that have been translated and summarized into English. For this task, we are using the GEM \cite{gehrmann-etal-2021-gem} version of the datasets, allowing us to make use of their fixes in the splitting of the datasets for training and testing.
\end{itemize}

These tasks allow us to explore summarization where the task involves documents and their summaries in the same language (MLSUM, XL-Sum), or where the task involves both translation and summarization at the same time (WikiLingua).

We note that with respect to task lengths, these multilingual tasks are not very long when compared to the tasks covered in the original LongT5 paper.
There is unfortunately a lack of lengthy, multilingual summarization tasks available, thus we use these three for comparisons.
As such, we tested with input lengths of 4k for input and 512 for output, which covers most documents for all the above tasks.

For all these tasks, we report standard ROUGE scores (ROUGE-1, ROUGE-2, and ROUGE-L).

\begin{table}[tb]
\small
\centering
\begin{tabular}{lccc}
\toprule
\multicolumn{4}{c}{\textbf{FR}} \\
\textbf{Approach} & R-1 & R-2 & R-L\\
\midrule
M-BERT & - & - & 25.09 \\
\midrule
mLongT5 (base) & 30.79 & 14.16 & 23.83 \\
mLongT5 (large) & 31.44 & 14.74 & 24.36 \\
mLongT5 (xl) & 32.18 & 15.68 & {\bf 25.18} \\
\toprule
\multicolumn{4}{c}{\textbf{DE}} \\
\textbf{Approach} & R-1 & R-2 & R-L \\
\midrule
M-BERT & - & - & 42.01 \\
\midrule
mLongT5 (base) & 45.60 & 35.31 & 42.22  \\
mLongT5 (large) & 46.21 & 35.68 & 42.71 \\
mLongT5 (xl) & 46.95 & 36.36 & {\bf 43.45} \\
\toprule
\multicolumn{4}{c}{\textbf{ES}} \\
\textbf{Approach} & R-1 & R-2 & R-L\\
\midrule
M-BERT & - & - & 20.44 \\
\midrule
mLongT5 (base) & 28.78 & 10.98 & 23.15 \\
mLongT5 (large) & 29.05 & 11.58 & 23.50 \\
mLongT5 (xl) & 30.36 & 12.77 & {\bf 24.73} \\
\toprule
\multicolumn{4}{c}{\textbf{TR}} \\
\textbf{Approach} & R-1 & R-2 & R-L\\
\midrule
M-BERT & - & - & 32.94 \\
\midrule
mLongT5 (base) & 44.18 & 30.86 & 38.60 \\
mLongT5 (large) & 44.92 & 31.55 & 39.29 \\
mLongT5 (xl) & 45.73 & 32.80 & {\bf 40.26} \\
\toprule
\multicolumn{4}{c}{\textbf{RU}} \\
\textbf{Approach} & R-1 & R-2 & R-L\\
\midrule
M-BERT & - & - & {\bf 9.48} \\
\midrule
mLongT5 (base) & 7.73 & 1.78 & 7.22  \\
mLongT5 (large) & 7.71 & 1.86 & 7.23 \\
mLongT5 (xl) & 8.85 & 2.67 & 8.42 \\
\bottomrule
\end{tabular}
\caption{MLSUM results comparing mLongT5 with the original model M-BERT.
Note that the original paper only reported ROUGE-L scores, while we also report ROUGE-1 and ROUGE-2.
}
\label{tab:mlsum_results}
\end{table}

\subsubsection{MLSUM}

Table \ref{tab:mlsum_results} shows our results for the MLSUM task.
We are comparing to the M-BERT \cite{devlin2018multilingual} model used in the original paper.
The authors only reported ROUGE-L scores, while we also report ROUGE-1 and ROUGE-2 scores.

Looking at the ROUGE-L scores, we can see that mLongT5 performs comparably to M-BERT for French, while doing better than M-BERT for all model sizes in German, Spanish, and Turkish.
It is only with Russian does it do slightly worse.
As noted in the original paper, Russian was the hardest language for language models, due to having a much smaller dataset when compared to the other languages in the corpus and a higher rate of novelty (words found in the summary but not in the input document).
Additionally, as we mentioned before, the dataset input lengths are not very long, thus models with full attention can take better advantage of the short lengths compared to mLongT5.
This can then contribute to mLongT5 not performing as well for this instance.

\begin{table*}[tb]
\small
\centering
\begin{tabular}{l|ccc|ccc|ccc|ccc}
\toprule
& \multicolumn{3}{c|}{\textbf{mT5 (base)}} & \multicolumn{3}{c|}{\textbf{mLongT5 (base)}} & \multicolumn{3}{c|}{\textbf{mLongT5 (large)}} & \multicolumn{3}{c}{\textbf{mLongT5 (xl)}} \\
\textbf{Language} & R-1 & R-2 & R-L & R-1 & R-2 & R-L  & R-1 & R-2 & R-L  & R-1 & R-2 & R-L  \\
\midrule
Gujarati & 21.96 & 7.74 & 19.86 & 19.59 & 6.08 & 17.61 & 22.38 & 7.94 & 20.15 & \textbf{25.52} & \textbf{9.92} & \textbf{22.78} \\
Marathi & 22.01 & 9.54 & 19.92 & 20.33 & 8.62 & 18.41 & 23.35 & 10.56 & 21.22 & \textbf{25.90} & \textbf{12.03} & \textbf{23.07} \\
Punjabi & 30.70 & 12.21 & 25.52 & 28.61 & 10.43 & 23.66 & 31.92 & 12.75 & 26.17 & \textbf{34.45} & \textbf{14.81} & \textbf{28.42} \\
Serbian \scriptsize{(Cyrillic)} & 23.78 & 7.98 & 20.14 & 20.30 & 5.86 & 16.74 & 21.92 & 6.98 & 18.35 & \textbf{27.51} & \textbf{11.46} & \textbf{23.49} \\
Serbian \scriptsize{(Latin)} & 21.64 & 6.66 & 18.23 & 18.14 & 4.75 & 14.96 & 21.79 & 6.92 & 18.14 & \textbf{25.86} & \textbf{10.17} & \textbf{21.76} \\
Vietnamese & 32.88 & 16.22 & 26.08 & 31.58 & 15.41 & 25.02 & 34.54 & 17.63 & 27.59 & \textbf{38.17} & \textbf{20.49} & \textbf{30.98} \\
\bottomrule
\end{tabular}
\caption{Results for XL-Sum, focusing on languages that have lengthier inputs. The rest of the results can be seen in the Appendix \ref{sec:app_xlsum_all}.}
\label{tab:xm_sum_results}
\end{table*}

\subsubsection{XL-Sum}

For XL-Sum, we finetuned the model in a similar approach to the original paper -- we finetuned on a mixture of all the languages for 50,000 steps, and then performed tests for each of the individual languages from this single model.

Table \ref{tab:xm_sum_results} shows a subset of the languages (the full results can be seen in Appendix \ref{sec:app_xlsum_all}).
We highlight languages that had longer input lengths (due to both the length of the original documents and how they are then subsequently tokenized by the SPM).

As we can see, mLongT5 performed well compared to mT5 for these lengthier inputs.
When comparing base to base, it did slightly worse, as expected with mT5 having full attention.
The original LongT5 model, when finetuned on datasets that are of shorter lengths, had also shown slightly worse performance when compared to a model of full attention.
We are seeing similar results here.
But mLongT5 is able to more easily scale to larger model sizes, and as such, we can see stronger results as we increase the size of the model.

\begin{table*}[tb]
\small
\centering
\begin{tabular}{l|ccc|ccc|ccc|ccc}
\toprule
 & \multicolumn{3}{c|}{\textbf{ES-EN}}  & \multicolumn{3}{c|}{\textbf{TR-EN}} & \multicolumn{3}{c|}{\textbf{RU-EN}} & \multicolumn{3}{c}{\textbf{VI-EN}} \\
\textbf{Approach} & R-1 & R-2 & R-L  & R-1 & R-2 & R-L  & R-1 & R-2 & R-L  & R-1 & R-2 & R-L \\
\midrule
mT5 (base) & 30.9 & 10.6 & 26.4  & 32.0 & 13.1 & 26.0  & 27.3 & 8.6 & 23.3 & 25.6 & 7.7 & 21.5\\
mT5 (large) & 34.2 & 12.6 & 29.1  & 34.0 & 14.5 & 27.5   & 32.3 & 11.2 & 26.9 & 32.1 & 10.9 & 26.0 \\
mT5 (xl) & {\bf 41.2} & 17.2 & {\bf 34.6} & 40.0 & 18.3 & 33.3  & 37.2 & 14.6 & 30.9 & 37.6 & 14.9 & 31.2 \\
\midrule
mLongT5 (base) & 36.1 & 14.0 & 30.3 & 34.5 & 14.9 & 28.6 & 32.4 & 11.6 & 26.5 & 32.3 & 11.7 & 26.4 \\
mLongT5 (large) & 38.2 & 15.5 & 32.0 & 38.1 & 17.5 & 32.0 & 34.4 & 13.1 & 28.5 & 35.1 & 13.8 & 29.1  \\
mLongT5 (xl) & 40.8 & {\bf 17.6} & 34.3 & {\bf 42.5} & {\bf 20.9} & {\bf 36.7} & {\bf 37.6} & {\bf 15.7} & {\bf 31.8} & {\bf 38.7} & {\bf 16.6} & {\bf 32.8}  \\
\bottomrule
\end{tabular}
\caption{WikiLingua summarization results.
These results are using the GEM version of the task.
}
\label{tab:wikilingua_results}
\end{table*}

\subsubsection{WikiLingua}

The final summarization task is WikiLingua, with results shown in Table \ref{tab:wikilingua_results}.
This task requires both translation and summarization, with the task translating from a full document of another language into an English summary.
As previously mentioned we are using the GEM version of this task, and compare our results to the mT5 model on their leaderboard.

As shown in the results, mLongT5 tends to do better for many of the model sizes across the 4 languages, with only slightly worse performance with XL size for Spanish.

\subsection{Question-Answering}
   
For question-answering, we applied mLongT5 to TyDi QA \cite{clark-etal-2020-tydi}.
TyDi QA is a multilingual task covering 11 languages, trying to answer questions given a Wikipedia article.
There are two versions of this task, and we focus on the Minimal Answer Span Task, in which one is trying to either find the minimal span that answer the question, give a yes/no answer if the question is a yes/no question, or Null if the question cannot be answered given the article.

Similar to the original LongT5 paper and their application to Natural Questions, we have re-defined this task from extracting answer spans to a seq2seq task of generating answer texts.
The results shown will then differ from the TyDi QA leaderboard.
As such, we have also run the similar mT5 model on the same task to get a baseline to compare against.
Additionally, as the test set is not available for this task, we use 90\% of the training data as the train set and remaining 10\% as the dev set, and use the original dev set as our test set for reporting metrics.

Unlike the summarization tasks, TyDi QA has much longer input lengths -- mean of 5,148 tokens and $90^{th}$ percentile of 12,967 tokens when tokenized with the SentencePiece model.
As such, for mT5 we tested with input lengths between 512 and 4k, while for mLongT5 we tested with input lengths between 4k and 16k.

Table \ref{tab:tydiqa_results} show the results of running mT5 and mLongT5 on this dataset.
For this task, we report metrics of Exact Match (EM) and F1 score.
As can be seen in the results, mLongT5 is able to better answer the questions given that it can handle longer input sequences.

\begin{table}[tb]
\small
\centering
\begin{tabular}{lcc}
\toprule
\textbf{Approach} & EM & F1\\
\midrule
mT5 (base - 512 input) & 37.16 & 49.99  \\
mT5 (base - 1k input) &  43.09 & 56.36  \\
mT5 (base - 2k input) &  44.63 & 58.12  \\
mT5 (base - 4k input) &  45.41 & 58.63  \\
mT5 (large - 512 input) & 40.96 & 54.08   \\
mT5 (large - 4k input) & 52.77 & 66.54 \\
mT5 (xl - 512 input) & 43.84 & 56.98 \\
mT5 (xl - 4k input) & 55.03 & 68.26   \\
\midrule
mLongT5 (base - 4k input) & 50.76 & 62.74  \\
mLongT5 (base - 8k input) & 51.21 & 63.66  \\
mLongT5 (base - 16k input) & 52.43 & 64.51  \\
mLongT5 (large - 4k input) & 54.04 & 66.75 \\
mLongT5 (large - 8k input) & 55.56 & 68.26  \\
mLongT5 (large - 16k input) & 55.93 & 68.66 \\
mLongT5 (xl - 4k input) & 58.52 & 70.86 \\
mLongT5 (xl - 8k input) & 59.6 & 71.86 \\
mLongT5 (xl - 16k input) & {\bf 60.42} & {\bf 72.63}  \\
\bottomrule
\end{tabular}
\caption{TyDi QA results.}
\label{tab:tydiqa_results}
\end{table}

\section{Conclusion}

We have presented our new model mLongT5.
It has the benefits of the efficient architecture of LongT5, with the ability to handle multingual inputs and outputs.
As our report shows, the model is able to perform well on a variety of summarization and question-answering tasks.

\section*{Limitations}

mLongT5 has the same limitations as seen in the original LongT5 model, in that they are more suited for tasks of lengthier inputs.
Tasks with shorter inputs will be better served by models like mT5 and umT5, which can take advantage of full attention.

\bibliography{custom}

\begin{thebibliography}{17}
\expandafter\ifx\csname natexlab\endcsname\relax\def\natexlab#1{#1}\fi

\bibitem[{Ainslie et~al.(2020)Ainslie, Onta\~{n}\'{o}n, Alberti, Cvicek,
  Fisher, Pham, Ravula, Sanghai, Wang, and Yang}]{ainslie2020etc}
Joshua Ainslie, Santiago Onta\~{n}\'{o}n, Chris Alberti, Vaclav Cvicek, Zachary
  Fisher, Philip Pham, Anirudh Ravula, Sumit Sanghai, Qifan Wang, and Li~Yang.
  2020.
\newblock \href {https://arxiv.org/abs/2004.08483} {{ETC}: Encoding long and
  structured inputs in transformers}.
\newblock \emph{arXiv preprint arXiv:2004.08483}.

\bibitem[{Beltagy et~al.(2020)Beltagy, Peters, and Cohan}]{beltagy-longformer}
Iz~Beltagy, Matthew~E. Peters, and Arman Cohan. 2020.
\newblock \href {https://doi.org/10.48550/ARXIV.2004.05150} {Longformer: The
  long-document transformer}.

\bibitem[{Chung et~al.(2023)Chung, Constant, Garcia, Roberts, Tay, Narang, and
  Firat}]{chung2023unimax}
Hyung~Won Chung, Noah Constant, Xavier Garcia, Adam Roberts, Yi~Tay, Sharan
  Narang, and Orhan Firat. 2023.
\newblock \href {http://arxiv.org/abs/2304.09151} {{UniMax}: Fairer and more
  effective language sampling for large-scale multilingual pretraining}.

\bibitem[{Clark et~al.(2020)Clark, Choi, Collins, Garrette, Kwiatkowski,
  Nikolaev, and Palomaki}]{clark-etal-2020-tydi}
Jonathan~H. Clark, Eunsol Choi, Michael Collins, Dan Garrette, Tom Kwiatkowski,
  Vitaly Nikolaev, and Jennimaria Palomaki. 2020.
\newblock \href {https://doi.org/10.1162/tacl_a_00317} {{T}y{D}i {QA}: A
  benchmark for information-seeking question answering in typologically diverse
  languages}.
\newblock \emph{Transactions of the Association for Computational Linguistics},
  8:454--470.

\bibitem[{Devlin(2018)}]{devlin2018multilingual}
Jacob Devlin. 2018.
\newblock {Multilingual BERT README}.
\newblock
  \url{https://github.com/google-research/bert/blob/master/multilingual.md}.

\bibitem[{Gehrmann et~al.(2021)Gehrmann, Adewumi, Aggarwal, Ammanamanchi,
  Aremu, Bosselut, Chandu, Clinciu, Das, Dhole, Du, Durmus, Du{\v{s}}ek,
  Emezue, Gangal, Garbacea, Hashimoto, Hou, Jernite, Jhamtani, Ji, Jolly, Kale,
  Kumar, Ladhak, Madaan, Maddela, Mahajan, Mahamood, Majumder, Martins,
  McMillan-Major, Mille, van Miltenburg, Nadeem, Narayan, Nikolaev,
  Niyongabo~Rubungo, Osei, Parikh, Perez-Beltrachini, Rao, Raunak, Rodriguez,
  Santhanam, Sedoc, Sellam, Shaikh, Shimorina, Sobrevilla~Cabezudo, Strobelt,
  Subramani, Xu, Yang, Yerukola, and Zhou}]{gehrmann-etal-2021-gem}
Sebastian Gehrmann, Tosin Adewumi, Karmanya Aggarwal, Pawan~Sasanka
  Ammanamanchi, Anuoluwapo Aremu, Antoine Bosselut, Khyathi~Raghavi Chandu,
  Miruna-Adriana Clinciu, Dipanjan Das, Kaustubh Dhole, Wanyu Du, Esin Durmus,
  Ond{\v{r}}ej Du{\v{s}}ek, Chris~Chinenye Emezue, Varun Gangal, Cristina
  Garbacea, Tatsunori Hashimoto, Yufang Hou, Yacine Jernite, Harsh Jhamtani,
  Yangfeng Ji, Shailza Jolly, Mihir Kale, Dhruv Kumar, Faisal Ladhak, Aman
  Madaan, Mounica Maddela, Khyati Mahajan, Saad Mahamood, Bodhisattwa~Prasad
  Majumder, Pedro~Henrique Martins, Angelina McMillan-Major, Simon Mille, Emiel
  van Miltenburg, Moin Nadeem, Shashi Narayan, Vitaly Nikolaev, Andre
  Niyongabo~Rubungo, Salomey Osei, Ankur Parikh, Laura Perez-Beltrachini,
  Niranjan~Ramesh Rao, Vikas Raunak, Juan~Diego Rodriguez, Sashank Santhanam,
  Jo{\~a}o Sedoc, Thibault Sellam, Samira Shaikh, Anastasia Shimorina,
  Marco~Antonio Sobrevilla~Cabezudo, Hendrik Strobelt, Nishant Subramani, Wei
  Xu, Diyi Yang, Akhila Yerukola, and Jiawei Zhou. 2021.
\newblock \href {https://doi.org/10.18653/v1/2021.gem-1.10} {The {GEM}
  benchmark: Natural language generation, its evaluation and metrics}.
\newblock In \emph{Proceedings of the 1st Workshop on Natural Language
  Generation, Evaluation, and Metrics (GEM 2021)}, pages 96--120, Online.
  Association for Computational Linguistics.

\bibitem[{Guo et~al.(2022)Guo, Ainslie, Uthus, Onta\~{n}\'{o}n, Ni, Sung, and
  Yang}]{guo-etal-2022-longt5}
Mandy Guo, Joshua Ainslie, David Uthus, Santiago Onta\~{n}\'{o}n, Jianmo Ni,
  Yun-Hsuan Sung, and Yinfei Yang. 2022.
\newblock \href {https://doi.org/10.18653/v1/2022.findings-naacl.55}
  {{L}ong{T}5: {E}fficient text-to-text transformer for long sequences}.
\newblock In \emph{Findings of the Association for Computational Linguistics:
  NAACL 2022}, pages 724--736, Seattle, United States. Association for
  Computational Linguistics.

\bibitem[{Hasan et~al.(2021)Hasan, Bhattacharjee, Islam, Mubasshir, Li, Kang,
  Rahman, and Shahriyar}]{hasan-etal-2021-xl}
Tahmid Hasan, Abhik Bhattacharjee, Md.~Saiful Islam, Kazi Mubasshir, Yuan-Fang
  Li, Yong-Bin Kang, M.~Sohel Rahman, and Rifat Shahriyar. 2021.
\newblock \href {https://doi.org/10.18653/v1/2021.findings-acl.413} {{XL}-sum:
  Large-scale multilingual abstractive summarization for 44 languages}.
\newblock In \emph{Findings of the Association for Computational Linguistics:
  ACL-IJCNLP 2021}, pages 4693--4703, Online. Association for Computational
  Linguistics.

\bibitem[{Kudo and Richardson(2018)}]{kudo-richardson-2018-sentencepiece}
Taku Kudo and John Richardson. 2018.
\newblock \href {https://doi.org/10.18653/v1/D18-2012} {{S}entence{P}iece: A
  simple and language independent subword tokenizer and detokenizer for neural
  text processing}.
\newblock In \emph{Proceedings of the 2018 Conference on Empirical Methods in
  Natural Language Processing: System Demonstrations}, pages 66--71, Brussels,
  Belgium. Association for Computational Linguistics.

\bibitem[{Ladhak et~al.(2020)Ladhak, Durmus, Cardie, and
  McKeown}]{ladhak-etal-2020-wikilingua}
Faisal Ladhak, Esin Durmus, Claire Cardie, and Kathleen McKeown. 2020.
\newblock \href {https://doi.org/10.18653/v1/2020.findings-emnlp.360}
  {{W}iki{L}ingua: A new benchmark dataset for cross-lingual abstractive
  summarization}.
\newblock In \emph{Findings of the Association for Computational Linguistics:
  EMNLP 2020}, pages 4034--4048, Online. Association for Computational
  Linguistics.

\bibitem[{Liu et~al.(2020)Liu, Gu, Goyal, Li, Edunov, Ghazvininejad, Lewis, and
  Zettlemoyer}]{liu-mbart}
Yinhan Liu, Jiatao Gu, Naman Goyal, Xian Li, Sergey Edunov, Marjan
  Ghazvininejad, Mike Lewis, and Luke Zettlemoyer. 2020.
\newblock \href {https://doi.org/10.1162/tacl_a_00343} {{Multilingual Denoising
  Pre-training for Neural Machine Translation}}.
\newblock \emph{Transactions of the Association for Computational Linguistics},
  8:726--742.

\bibitem[{Raffel et~al.(2019)Raffel, Shazeer, Roberts, Lee, Narang, Matena,
  Zhou, Li, and Liu}]{t5}
Colin Raffel, Noam Shazeer, Adam Roberts, Katherine Lee, Sharan Narang, Michael
  Matena, Yanqi Zhou, Wei Li, and Peter~J. Liu. 2019.
\newblock \href {http://arxiv.org/abs/1910.10683} {Exploring the limits of
  transfer learning with a unified text-to-text transformer}.
\newblock \emph{CoRR}, abs/1910.10683.

\bibitem[{Scialom et~al.(2020)Scialom, Dray, Lamprier, Piwowarski, and
  Staiano}]{scialom-etal-2020-mlsum}
Thomas Scialom, Paul-Alexis Dray, Sylvain Lamprier, Benjamin Piwowarski, and
  Jacopo Staiano. 2020.
\newblock \href {https://doi.org/10.18653/v1/2020.emnlp-main.647} {{MLSUM}: The
  multilingual summarization corpus}.
\newblock In \emph{Proceedings of the 2020 Conference on Empirical Methods in
  Natural Language Processing (EMNLP)}, pages 8051--8067, Online. Association
  for Computational Linguistics.

\bibitem[{Tay et~al.(2022)Tay, Dehghani, Tran, Garcia, Wei, Wang, Chung, Bahri,
  Schuster, Zheng, Zhou, Houlsby, and Metzler}]{ul2}
Yi~Tay, Mostafa Dehghani, Vinh~Q. Tran, Xavier Garcia, Jason Wei, Xuezhi Wang,
  Hyung~Won Chung, Dara Bahri, Tal Schuster, Huaixiu~Steven Zheng, Denny Zhou,
  Neil Houlsby, and Donald Metzler. 2022.
\newblock \href {https://doi.org/10.48550/ARXIV.2205.05131} {{UL2}: Unifying
  language learning paradigms}.

\bibitem[{Xue et~al.(2021)Xue, Constant, Roberts, Kale, Al-Rfou, Siddhant,
  Barua, and Raffel}]{xue-etal-2021-mt5}
Linting Xue, Noah Constant, Adam Roberts, Mihir Kale, Rami Al-Rfou, Aditya
  Siddhant, Aditya Barua, and Colin Raffel. 2021.
\newblock \href {https://doi.org/10.18653/v1/2021.naacl-main.41} {m{T}5: A
  massively multilingual pre-trained text-to-text transformer}.
\newblock In \emph{Proceedings of the 2021 Conference of the North American
  Chapter of the Association for Computational Linguistics: Human Language
  Technologies}, pages 483--498, Online. Association for Computational
  Linguistics.

\bibitem[{Zaheer et~al.(2020)Zaheer, Guruganesh, Dubey, Ainslie, Alberti,
  Ontanon, Pham, Ravula, Wang, Yang, and Ahmed}]{zaheer-bigbird}
Manzil Zaheer, Guru Guruganesh, Kumar~Avinava Dubey, Joshua Ainslie, Chris
  Alberti, Santiago Ontanon, Philip Pham, Anirudh Ravula, Qifan Wang, Li~Yang,
  and Amr Ahmed. 2020.
\newblock \href
  {https://proceedings.neurips.cc/paper/2020/file/c8512d142a2d849725f31a9a7a361ab9-Paper.pdf}
  {{Big Bird}: Transformers for longer sequences}.
\newblock In \emph{Advances in Neural Information Processing Systems},
  volume~33, pages 17283--17297. Curran Associates, Inc.

\bibitem[{Zhang et~al.(2020)Zhang, Zhao, Saleh, and Liu}]{pegasus}
Jingqing Zhang, Yao Zhao, Mohammad Saleh, and Peter~J. Liu. 2020.
\newblock \href {https://proceedings.mlr.press/v119/zhang20ae.html} {{PEGASUS}:
  Pre-training with extracted gap-sentences for abstractive summarization}.
\newblock In \emph{Proceedings of the 37th International Conference on Machine
  Learning}, volume 119 of \emph{Proceedings of Machine Learning Research},
  pages 11328--11339. PMLR.

\end{thebibliography}
\bibliographystyle{acl_natbib}

\appendix

\section{XL-Sum}\label{sec:app_xlsum_all}

\begin{table*}
\small
\centering
\begin{tabular}{l|ccc|ccc|ccc|ccc}
\toprule
& \multicolumn{3}{c|}{\textbf{mT5 (base)}} & \multicolumn{3}{c|}{\textbf{mLongT5 (base)}} & \multicolumn{3}{c|}{\textbf{mLongT5 (large)}} & \multicolumn{3}{c}{\textbf{mLongT5 (xl)}} \\
\textbf{Language} & R-1 & R-2 & R-L & R-1 & R-2 & R-L  & R-1 & R-2 & R-L  & R-1 & R-2 & R-L  \\
\midrule
Amharic & 20.05 & 7.41 & 18.08 & 16.70 & 5.91 & 14.73 & 20.29 & 7.99 & 18.09 & 22.37 & 8.90 & 19.91 \\
Arabic & 34.91 & 14.79 & 29.16  & 26.39 & 11.01 & 22.45 & 27.65 & 12.25 & 23.57 & 32.09 & 15.04 & 27.74 \\
Azerbaijani & 21.42 & 9.52 & 19.33 & 17.52 & 7.10 & 15.77 & 19.92 & 8.80 & 18.08 & 22.68 & 9.89 & 20.36 \\
Bengali & 29.57 & 12.11 & 25.13 & 21.39 & 8.22 & 18.65 & 24.69 & 10.04 & 21.25 & 26.83 & 11.32 & 22.86 \\
Burmese & 15.96 & 5.15 & 14.18 & 45.28 & 26.62 & 34.76 & 49.07 & 29.52 & 38.10 & 51.60 & 31.69 & 40.20 \\
Chinese (Simp.) & 39.41 & 17.79 & 33.41 & 38.90 & 21.78 & 32.59 & 42.62 & 24.70 & 35.80 & 48.42 & 29.99 & 41.28 \\
Chinese (Trad.) & 37.19 & 17.14 & 31.62 & 39.45 & 22.40 & 32.51 & 43.32 & 25.56 & 35.95 & 48.82 & 30.80 & 41.18 \\
English & 37.60 & 15.15 & 29.88 & 32.85 & 11.38 & 25.64 & 35.59 & 13.63 & 28.02 & 39.51 & 17.00 & 31.77 \\
French & 35.34 & 16.17 & 28.20 & 30.06 & 12.93 & 24.21 & 31.88 & 14.32 & 25.61 & 34.82 & 16.17 & 28.11 \\
Gujarati & 21.96 & 7.74 & 19.86 & 19.59 & 6.08 & 17.61 & 22.38 & 7.94 & 20.15 & 25.52 & 9.92 & 22.78 \\
Hausa & 39.44 & 17.68 & 31.67 & 34.61 & 13.73 & 27.30 & 38.04 & 16.07 & 30.32 & 40.58 & 18.57 & 32.52 \\
Hindi & 38.59 & 16.88 & 32.01  & 34.81 & 14.29 & 28.71 & 37.42 & 16.71 & 31.22 & 40.92 & 19.73 & 34.41 \\
Igbo & 31.61 & 10.16 & 24.53 & 25.82 & 8.05 & 20.19 & 30.41 & 10.01 & 23.68 & 31.31 & 9.88 & 24.07 \\
Indonesian & 37.00 & 17.02 & 30.76 & 32.15 & 13.05 & 26.59 & 35.17 & 15.23 & 29.07 & 38.87 & 18.00 & 32.64 \\
Japanese & 48.15 & 23.85 & 37.36 & 45.56 & 27.12 & 36.51 & 48.60 & 29.95 & 39.00 & 50.77 & 32.06 & 40.79 \\
Kirundi & 31.99 & 14.37 & 25.83 & 25.61 & 10.07 & 20.26 & 29.36 & 12.78 & 23.67 & 31.67 & 14.55 & 25.50 \\
Korean & 23.67 & 11.45 & 22.36 & 20.25 & 9.20 & 19.00 & 23.18 & 10.42 & 21.38 & 25.30 & 11.63 & 23.31 \\
Kyrgyz & 18.38 & 7.96 & 16.50  & 14.08 & 5.27 & 12.46 & 16.01 & 6.30 & 14.14 & 18.19 & 7.81 & 16.00 \\
Marathi & 22.01 & 9.54 & 19.92 & 20.33 & 8.62 & 18.41 & 23.35 & 10.56 & 21.22 & 25.90 & 12.03 & 23.07 \\
Nepali & 26.65 & 10.25 & 24.28 & 23.96 & 8.94 & 21.80 & 26.24 & 10.33 & 23.91 & 28.87 & 11.59 & 26.17 \\
Oromo & 18.70 & 6.17 & 16.19 & 14.88 & 4.38 & 12.71 & 17.91 & 5.65 & 15.28 & 19.52 & 6.50 & 17.18 \\
Pashto & 38.47 & 15.55 & 31.91 & 35.01 & 13.79 & 28.84 & 38.63 & 16.06 & 32.00 & 41.37 & 17.61 & 33.92 \\
Persian & 36.94 & 16.19 & 30.07 & 35.47 & 14.66 & 28.40 & 37.70 & 16.45 & 30.49 & 40.64 & 18.89 & 33.16 \\
Pidgin & 37.96 & 15.12 & 29.87 & 33.86 & 12.01 & 26.68 & 35.86 & 13.72 & 28.24 & 38.01 & 15.08 & 29.78 \\
Portuguese & 37.17 & 15.90 & 28.56 & 31.67 & 12.51 & 24.46 & 34.04 & 14.51 & 26.65 & 37.66 & 17.57 & 29.88 \\
Punjabi & 30.70 & 12.21 & 25.52 & 28.61 & 10.43 & 23.66 & 31.92 & 12.75 & 26.17 & 34.45 & 14.81 & 28.42 \\
Russian & 32.22 & 13.64 & 26.17  & 22.11 & 8.29 & 18.62 & 24.39 & 10.00 & 20.54 & 28.20 & 12.72 & 23.91 \\
Scottish Gaelic & 29.02 & 10.99 & 22.88 & 26.98 & 8.87 & 21.57 & 29.80 & 10.64 & 23.44 & 31.74 & 12.61 & 25.65 \\
Serbian (Cyrillic) & 23.78 & 7.98 & 20.14 & 20.30 & 5.86 & 16.74 & 21.92 & 6.98 & 18.35 & 27.51 & 11.46 & 23.49 \\
Serbian (Latin) & 21.64 & 6.66 & 18.23 & 18.14 & 4.75 & 14.96 & 21.79 & 6.92 & 18.14 & 25.86 & 10.17 & 21.76 \\
Sinhala & 27.29 & 13.38 & 23.47 & 22.69 & 10.02 & 19.96 & 25.24 & 11.52 & 21.98 & 27.78 & 13.20 & 24.45 \\
Somali & 31.56 & 11.58 & 24.22  & 27.85 & 9.08 & 21.10 & 30.29 & 10.69 & 23.29 & 31.64 & 11.11 & 24.28 \\
Spanish & 31.51 & 11.88 & 24.07 & 26.82 & 9.05 & 20.47 & 28.71 & 10.56 & 22.04 & 32.20 & 13.10 & 24.88 \\
Swahili & 37.67 & 17.85 & 30.91  & 31.79 & 13.25 & 25.67 & 34.29 & 15.22 & 27.82 & 37.29 & 17.22 & 30.96 \\
Tamil & 24.33 & 11.06 & 22.07  & 20.68 & 8.67 & 18.71 & 24.08 & 10.74 & 21.71 & 26.81 & 12.23 & 24.21 \\
Telugu & 19.86 & 7.03 & 17.61 & 15.11 & 4.69 & 13.48 & 17.98 & 6.12 & 16.10 & 21.20 & 7.77 & 18.88 \\
Thai & 37.40 & 17.28 & 28.88 & 35.98 & 21.39 & 26.65 & 38.11 & 22.92 & 28.26 & 40.70 & 25.23 & 30.12 \\
Tigrinya & 25.32 & 8.02 & 21.17 & 22.27 & 7.08 & 18.61 & 26.30 & 8.90 & 22.05 & 28.53 & 10.13 & 24.05 \\
Turkish & 32.93 & 15.57 & 29.26 & 25.52 & 11.54 & 22.83 & 28.56 & 13.62 & 25.72 & 31.33 & 15.61 & 28.20 \\
Ukrainian & 23.99 & 10.14 & 20.92 & 20.97 & 8.16 & 18.17 & 23.34 & 9.74 & 20.29 & 27.05 & 12.16 & 23.68 \\
Urdu & 39.56 & 18.37 & 32.84 & 37.11 & 15.97 & 30.14 & 39.90 & 18.53 & 32.75 & 43.03 & 21.40 & 35.72 \\
Uzbek & 16.83 & 6.34 & 15.41 & 14.60 & 5.36 & 13.39 & 17.26 & 6.42 & 15.49 & 19.18 & 7.80 & 17.29 \\
Vietnamese & 32.88 & 16.22 & 26.08 & 31.58 & 15.41 & 25.02 & 34.54 & 17.63 & 27.59 & 38.17 & 20.49 & 30.98 \\
Welsh & 32.66 & 11.60 & 26.12 & 29.96 & 9.40 & 23.96 & 33.66 & 12.26 & 27.01 & 36.49 & 15.34 & 29.79 \\
Yoruba & 31.66 & 11.66 & 25.09 & 25.87 & 8.99 & 20.27 & 29.49 & 10.50 & 23.26 & 32.20 & 12.34 & 25.84 \\
\bottomrule
\end{tabular}
\caption{Full results for XL-Sum.}
\label{tab:xm_sum_all_results}
\end{table*}

We show the full results of running our mLongT5 models on XL-Sum in Table \ref{tab:xm_sum_all_results}.
These results are those that had been uploaded to GitHub \footnote{\url{https://github.com/csebuetnlp/xl-sum}} by the authors along with the updated datasets.

When computing ROUGE scores, we use similar computations as done in the respective paper, with exceptions to Chinese, Japanese and Thai.
For these languages, we use the SPM we used in our model for the tokenization of the results in order to compute ROUGE.

\end{document}